\documentclass{scrartcl}
\usepackage{graphicx}
\usepackage{tabularray}
\usepackage{makecell}
\usepackage{multirow}
\usepackage{booktabs}
\usepackage{siunitx}
\usepackage{caption}    
\usepackage{subcaption} 
\usepackage{amssymb}    
\usepackage{appendix}   
\usepackage{dramatist}  
\usepackage{parskip}    
\usepackage[hidelinks]{hyperref}
\usepackage[authordate,backend=biber,ibidtracker=false]{biblatex-chicago}
\title{\textbf{From stage to page: language independent bootstrap measures of distinctiveness in fictional speech}}
\date{}

\begin{document}
\maketitle

\begin{center}
\begin{tblr}{|Q[c,0.49\linewidth]|Q[c,0.49\linewidth]|}

 \hline
  \thead{\textbf{Artjoms Šeļa} \\  Institute of Polish Language, \\  Polish Academy of Sciences \\ and University of Tartu \\ artjoms.sela@ijp.pan.pl} & \thead{\textbf{Ben Nagy} \\ Institute of Polish Language, \\ Polish Academy of Sciences \\ benjamin.nagy@ijp.pan.pl} \\
\hline
  \thead{\textbf{Joanna Byszuk} \\  Institute of Polish Language, \\ Polish Academy of Sciences \\ joanna.byszuk@ijp.pan.pl} &  \thead{\textbf{Laura Hernández-Lorenzo} \\ University of Seville \\ lhernandez1@us.es} \\
  \hline
 \thead{\textbf{Botond Szemes} \\Research Centre for the Humanities \\ Institute for Literary Studies Budapest
 \\szemes.botond@abtk.hu} & \thead{\textbf{Maciej Eder} \\  Institute of Polish Language,\\ Polish Academy of Sciences \\ maciej.eder@ijp.pan.pl} \\
  \hline
  
\end{tblr}
\end{center}

\vspace{0.5cm}

{\footnotesize
\begin{center}
\textbf{Abstract}
\end{center}
\begin{changemargin}{3cm}{3cm}
Stylometry is mostly applied to \emph{authorial} style. More recently, researchers have begun investigating the style of \emph{characters}, finding that, although there is detectable stylistic variation, the variation remains within authorial bounds. In this article, we address the stylistic distinctiveness of characters in drama. Our primary contribution is methodological; we introduce and evaluate two non-parametric methods to produce a summary statistic for character distinctiveness that can be usefully applied and compared across languages and times. This is a significant advance---previous approaches have either been based on pairwise similarities (which cannot be easily compared) or indirect methods that attempt to infer distinctiveness using classification accuracy. Our first method is based on bootstrap distances between 3-gram probability distributions, the second (reminiscent of `unmasking' techniques) on word keyness curves. Both methods are validated and explored by applying them to a reasonably large corpus (a subset of DraCor): we analyse 3301 characters drawn from 2324 works, covering five centuries and four languages (French, German, Russian, and the works of Shakespeare). Both methods appear useful; the 3-gram method is statistically more powerful but the word keyness method offers rich interpretability. Both methods are able to capture phonological differences such as accent or dialect, as well as broad differences in topic and lexical richness. Based on exploratory analysis, we find that smaller characters tend to be more distinctive, and that women are cross-linguistically more distinctive than men, with this latter finding carefully interrogated using multiple regression. This greater distinctiveness stems from a historical tendency for female characters to be restricted to an `internal narrative domain' covering mainly direct discourse and family/romantic themes. It is hoped that direct, comparable statistical measures will form a basis for more sophisticated future studies, and advances in theory.

\end{changemargin}
}

\section{Introduction}

Since Vladimir Propp's work, structural narratology has approached fictional characters mainly through their role or function---by what they do, or what is done to them \parencite{eder_characters_2010}. This character typology relied on recurring functions in the narrative (lover, villain, victim, detective, etc.) and the same perspective was often adopted in computational research, where characters in novels were modelled on the basis of narrative passages rather than dialogue \parencite{bamman_bayesian_2014,bonch-osmolovskaya_text_2017,underwood_transformation_2018,stammbach_heroes_2022}.

In dramatic texts, however, the dominant device for characterisation is an utterance. While the script usually contains some stage directions, the specifics of characterisation and style of performance are not determined by the text itself, but developed by a specific theatre, director or a troupe. Over the course of history, many plays were written for specific theatre stages, and it was common practice to write characters for specific actors \parencite{fischer-lichte_history_2002}. Of course, this kind of `outsourced characterisation' was supported by dramatic conventions and formulas. Viewers’ expectations could be shaped without a single word being uttered on stage, just by a character wearing a costume, operating a puppet, or changing a dell'arte stock mask. At the same time, the things characters say and how they say them are the main textual source of information about them. It is reasonable to assume that dramatists make significant efforts to create linguistic distinctions between princes and paupers, lovers and schemers, aristocrats and merchants. Tragic monologue is written differently to a comedic exchange between servants. Some previous computational works treat linguistic distinctiveness of characters from the perspective of this stylistic continuum \parencite{vishnubhotla_are_2019}, noting that it can be influenced by genre, character gender, or their social and professional dispositions.

A parallel narratological tradition, tied to Bakhtin’s ideas of heteroglossia, focuses not on abstract character roles, but on the words characters say \parencite{bronwen_fictional_2012,culpeper_language_2001,sternberg_proteus_1982}. The modern novelistic space of dialogic exchange, `educated conversation' \parencite{moretti_network_2011} and the clash of styles in reported discourse become central here. Available stylometric research on fictional speech and micro-stylistic variation suggests that characters within a text are often distinguishable by their local linguistic patterns without obscuring the global authorial trace \parencite{burrows_computation_1987,hoover_microanalysis_2017}. As put by Burrows and Craig: `Characters speak in measurably different ways, but the authorial contrasts transcend this differentiation. The diversity of styles within an author always remains within bounds' \parencite[307--8]{burrows_authors_2012}.

Conceptually and methodologically, the majority of previous works examined not the \emph{distinctiveness} of characters, but their (pairwise) \emph{similarity}. Similarity measures are meaningful in pairwise contexts, but cannot be analysed and compared as individual summary statistics. Since Burrows' seminal study of speech patterns in Jane Austen’s characters \parencite{burrows_computation_1987}, these approaches focused on calculating similarity within a collection of characters: how different is character X from character Y, and each of them from character Z. Burrows measured the correlation between characters’ usage of 30 most frequent words (technically, he fit a linear regression for two sets of log-frequencies); later, similarity was most often inferred through clustering based on pairwise distance calculations \parencite{hoover_microanalysis_2017, reeve_imperial_2015, craig_style_2017}. Sometimes linguistic similarity served as a basis for arguing functional similarity, as well. A recent study that linked Bakhtin’s dialogism and the stylistic diversity of characters’ speech \parencite{vishnubhotla_are_2019}, proposed the analysis of distinctiveness rather than similarity using supervised classification. Instead of using a network of pairwise relationships, the authors asked how well a classifier can recognise character X as being written by author A. Classification accuracy in this scenario becomes an explicit summary statistic for distinctiveness that can be assigned to a character (or, in an aggregated manner, to a play or an author). However, the supervised approach, proposed by Vishnubhotla et al., is data hungry: it suffers from extreme class imbalance, an abundance of short samples (most characters speak only a little) and is dependent on language-specific feature construction procedures.

By contrast, this paper will present a simple, non-parametric measure of character distinctiveness that is based on bootstrapped probability distributions representing a character and all others present in a given play: an approach largely informed by authorship verification techniques. This measure is language-independent and relies only on the context of a single work, which, in turn, minimises problems of language variation, authorial signal and chronological change in a comparative setting. Individual distinctiveness scores can be then tested against other measures and metadata categories in a hypothesis-driven manner, not only across languages, but also across genres (e.g. novel vs. drama). Do comedies tend to employ more distinct characters? Does distinctiveness increase (authors get better), or decrease (social and linguistic homogenisation occurs) over time?  Is there a difference between the distinctiveness of fictional women and men? If so is it the direct result of perceived gender differences, or is it constructed by imagined differences in social and professional status? 

Lacking good descriptive metadata on the dramatic characters, this paper will not answer above mentioned questions in any satisfying way. Instead we focus on presenting and justifying the measure of distinctiveness and exploring several factors that might shape the final scores (like the year of composition, character gender and characters’ sample size). 

\section{Materials}

\begin{table}[h]
\centering
\begin{tabular}{lrrrrrr}
            &       Total & Characters & Unique & Unique & Total & Total \\
     Corpus &       Characters &  Analysed &  3-grams &  Words &  3-grams &  Words \\
\midrule
     French &        15462 &        1744 &            9896 &         79994 &       29.79 m &      5.47 m \\
     German &        14010 &        1182 &           14341 &        150956 &       24.80 m &      4.31 m \\
    Russian &         3707 &         248 &           12542 &         71217 &        4.05 m &      0.72 m \\
Shakespeare &         1431 &         127 &            5921 &         19595 &        2.16 m &      0.43 m \\

\end{tabular}
\caption{A summary of the corpus. All word and 3-gram counts are for the filtered corpus (characters that speak at least 2000 words) only.}
\label{tab:corpus}
\end{table}

As the beginning of our exploration of cross-linguistic variation, we examined four dramatic corpora from DraCor \parencite{fischer_programmable_2019}: Shakespeare, French, German, and Russian. DraCor is a project that gathers dramatic corpora in various languages, primarily European, encoded in TEI-XML. With 15 corpora available so far, including the Shakespeare corpus available both in English and German, DraCor facilitates large scale analysis of dramatic conventions across language traditions, and offers a wide variety of useful metadata, at the level of both plays and characters. While the analysis of all DraCor corpora would be possible with the methods we developed, for the purpose of this preliminary study we focused on the languages and dramatic traditions well known to the members of our team, eventually selecting the full corpora for Shakespeare, French, German, and Russian: a total of 2324 texts, the majority of which come from French and German. The corpus is summarised in Table \ref{tab:corpus}.

\section{Methods}

\subsection{General Approach and Definitions}

\begin{figure}
  \centering
  \includegraphics[width=0.95\textwidth]{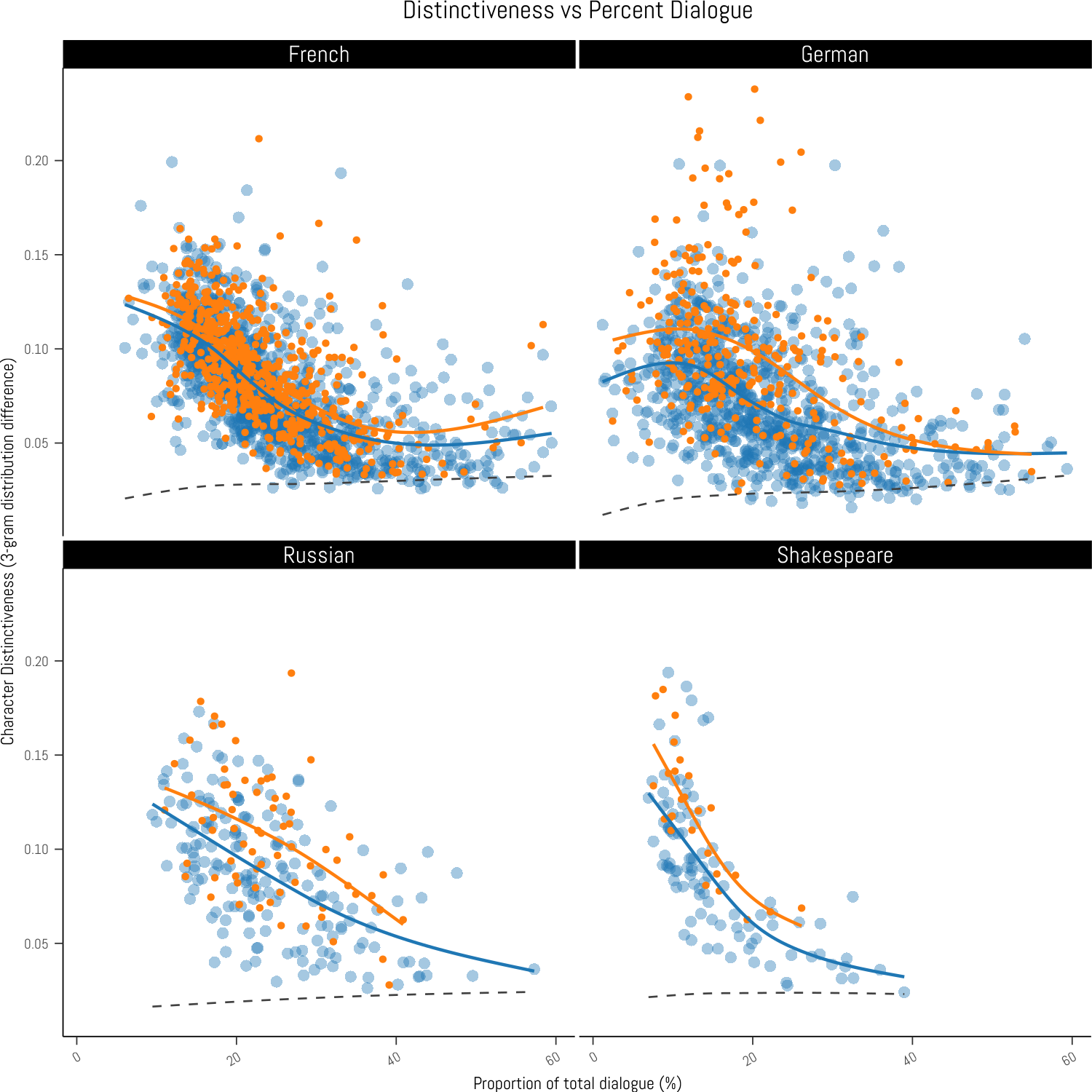}
  \caption{Character distinctiveness, per corpus, versus \% Dialogue. Women are shown smaller, in orange, men (and undefined) larger and in blue. GAM (Generalised Additive Model) trendlines are superimposed in the same colours. Baseline data (GAM trend for distinctiveness of character vs self) is shown as a dashed line.}
  \label{fig:distinctiveness_dialog}
\end{figure}

\begin{figure}
  \centering
  \includegraphics[width=0.95\textwidth]{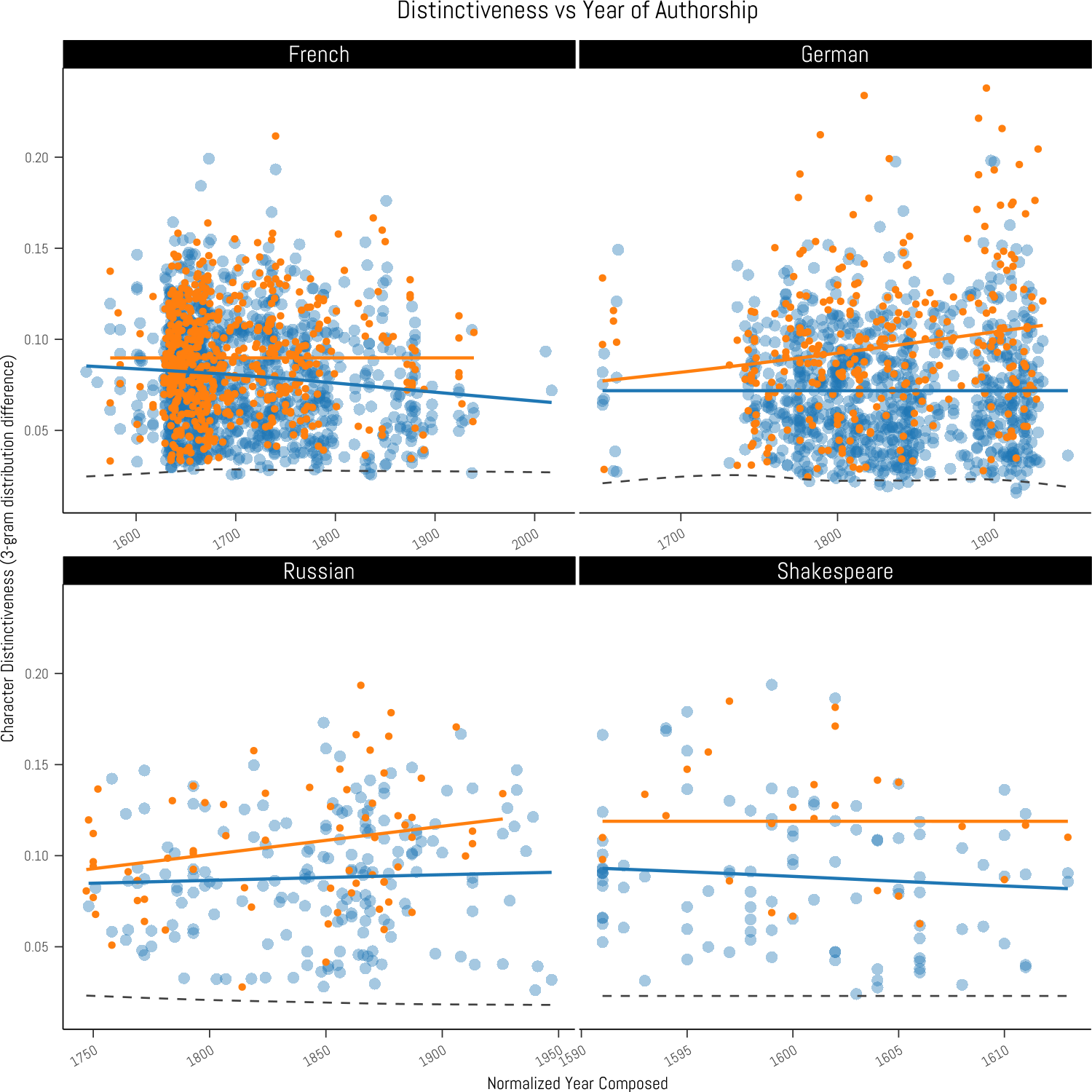}
  \caption{Character distinctiveness, per corpus, versus year composed (DraCor data). Women are shown smaller, in orange, men (and undefined) larger and in blue. GAM (Generalised Additive Model) trendlines are superimposed in the same colours. Baseline data (GAM trend for distinctiveness of character vs self) is shown as a dashed line.}
  \label{fig:distinctiveness_year}
\end{figure}

\begin{figure}
\centering
\begin{subfigure}[c]{0.3\textwidth}
  \centering
  \includegraphics[width=\textwidth]{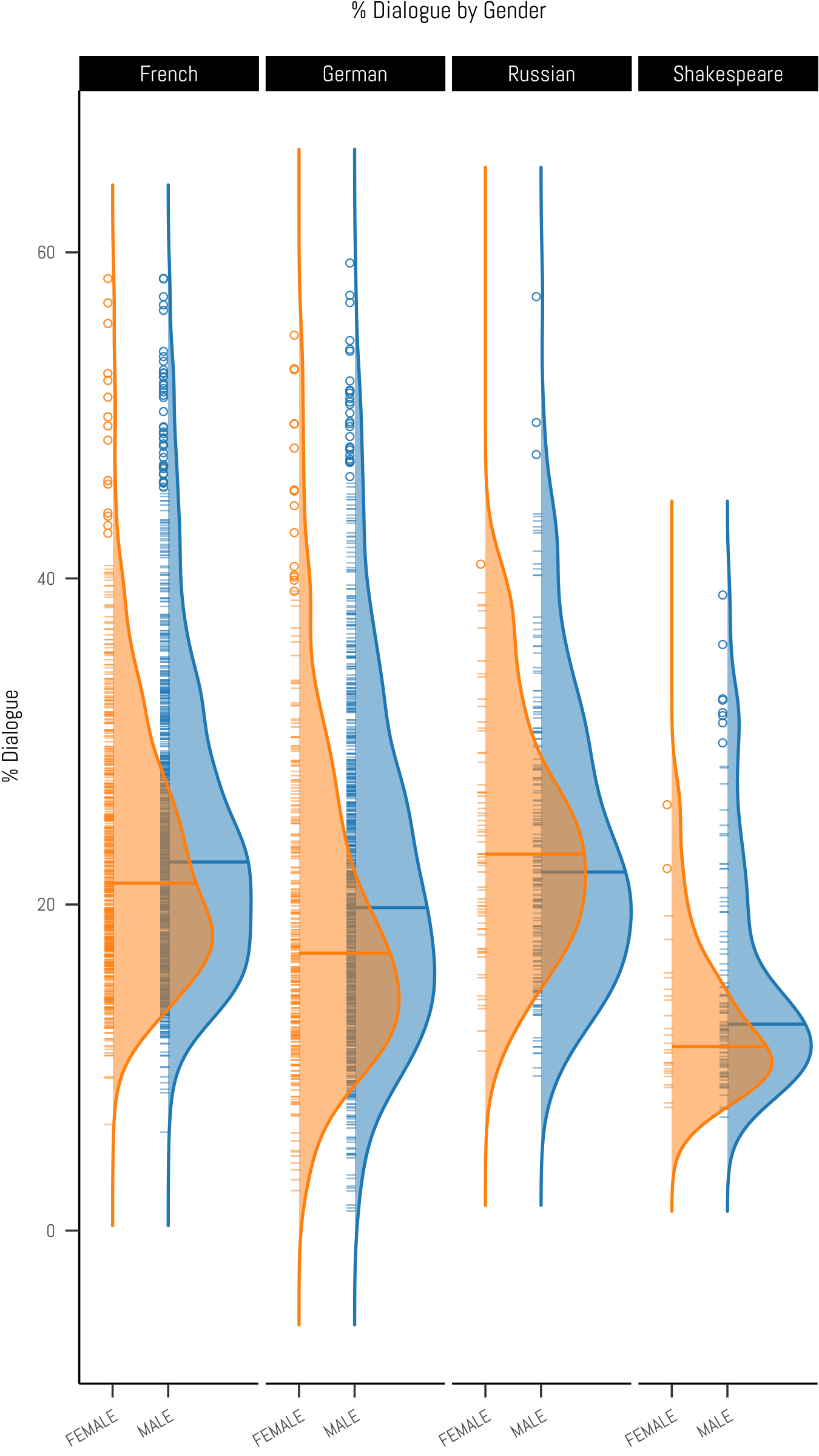}
  \caption{\% Dialogue}
  \label{fig:dialog_gender}
\end{subfigure}
\begin{subfigure}[c]{0.3\textwidth}
  \centering
  \includegraphics[width=\textwidth]{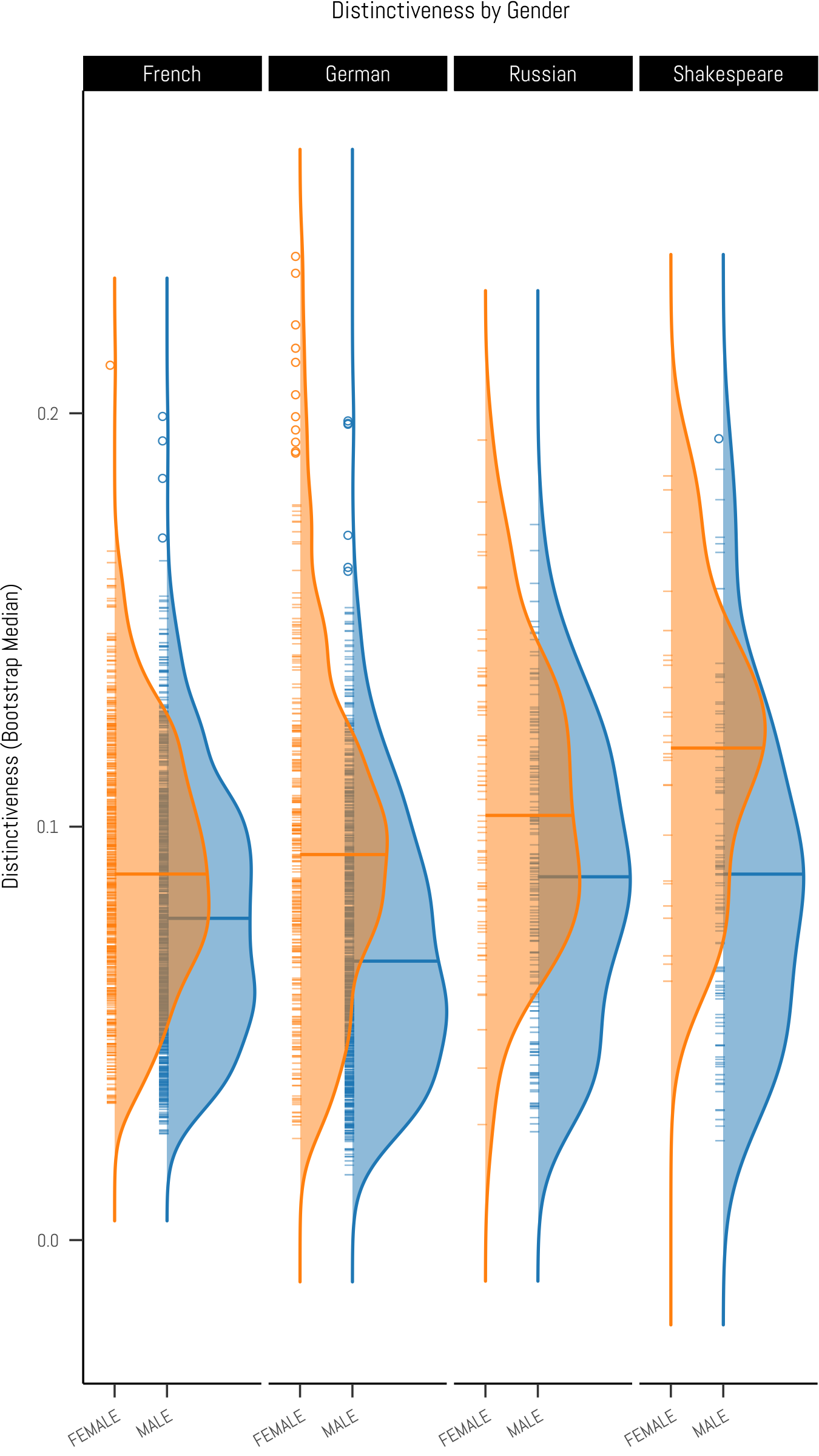}
  \caption{Distinctiveness}
  \label{fig:distinctiveness_gender}
\end{subfigure}
\begin{subfigure}[c]{0.3\textwidth}
  \centering
  \includegraphics[width=\textwidth]{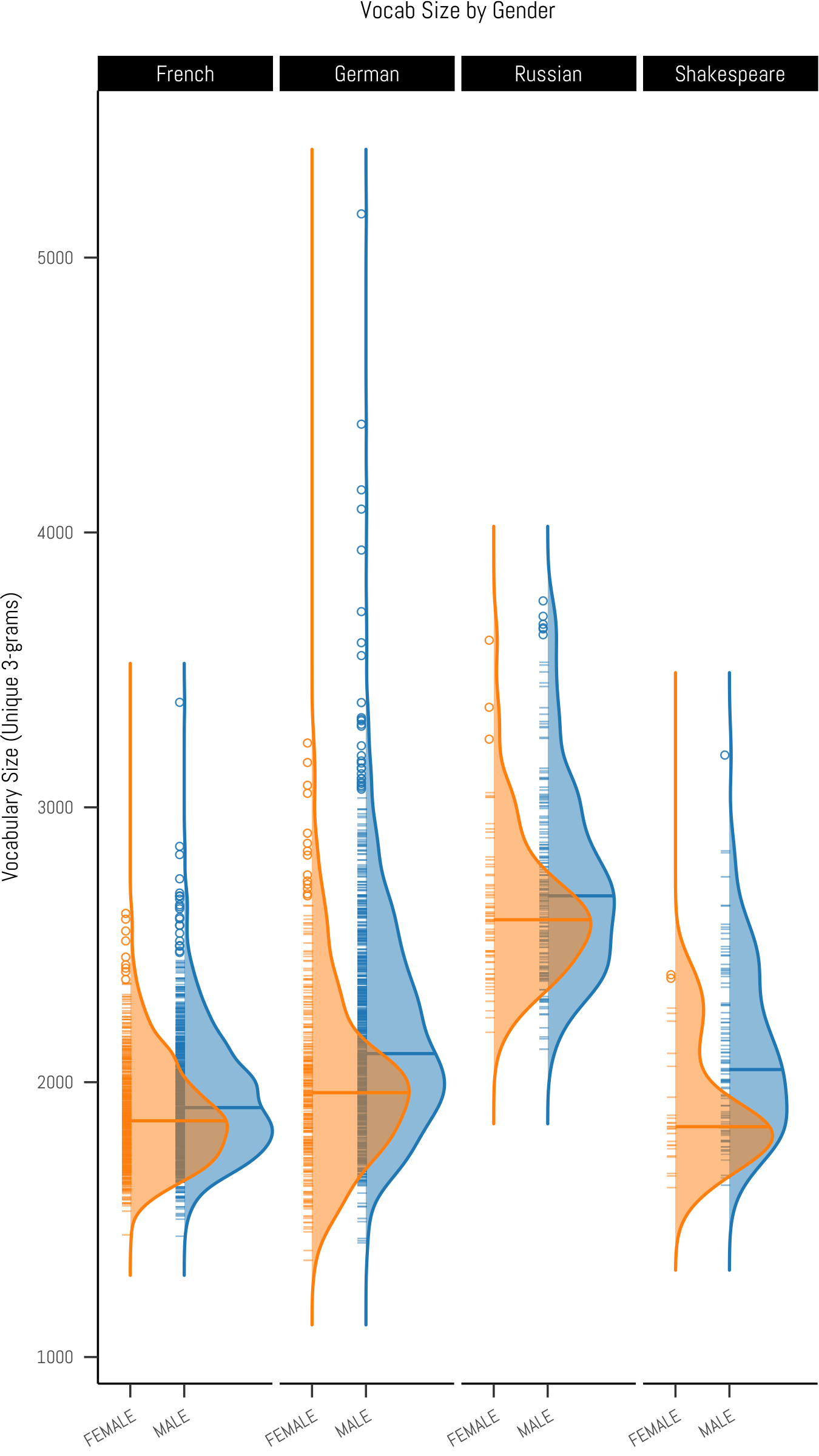}
  \caption{Vocab size (3-grams)}
  \label{fig:vocab_gender}
\end{subfigure}
\caption{An analysis, per corpus, of the distribution of various features by gender. Distributions are estimated, with the median shown as a solid line. Actual points are shown as rug plots with outliers `o' plotted for points outside 3Q + 2$\times$IQR.}
\label{fig:gender}
\end{figure}

Our understanding of character distinctiveness is largely informed by `authorship verification' approaches, which centre around verifying that a text is written by a target author. This problem is more general than `authorship attribution' that tries to identify the nearest stylistic neighbour for a text \parencite{halvani_assessing_2019}. Instead, authorship verification asks about the relative \emph{magnitude} of similarity: is a target text more similar to same-author samples, or different-author samples? With this in mind, we define a character's `distinctiveness' as the degree to which the style of their speech differs from that of other characters. We understand `style' here instrumentally, as a deviation from an unobserved average language \parencite{herrmann_revisiting_2015} and do not introduce aggressive feature filtering, allowing both `grammatical' and `thematic' signal to contribute to the final measures. We anchor our distinctiveness measure in the context of the specific text in which a character appears. In theory, the frame of reference could be all plays from one author, or all plays from the same period, or even some external corpus---however all of these would greatly complicate any comparative study.

\subsection{Bootstrap 3-gram Distinctiveness}

Based on our definition of distinctiveness above, we considered a character’s style to be an idiolect sampled from a frequency distribution of character 3-grams. As a natural language distribution, this was expected to be generally Zipfian, a family of heavy-tailed distributions, so non-parametric methods were seen to be important. 3-grams were preferred to words for a number of reasons: first, they capture sub-word information which means they will reflect general sonic preferences (so they can capture things like accent) and, particularly in inflected languages, also reflect some grammatical style; second, as a practical matter, they effectively expand the sample data, since a string of text produces approximately one 3-gram per character. This increased sample size should reduce the variance of the statistics. Finally, the number of unique 3-grams in a language is considerably smaller than the number of words, so the frequency data is less sparse, which again is expected to increase robustness. To now operationalise the distinctiveness, as defined, we used standard bootstrap methods to measure the median energy distance \parencite{szekely_energy_2013} with bootstrap confidence intervals between the two distributions (character 3-gram frequencies vs `other’ 3-gram frequencies). The energy distance is one of a family of related metrics that are commonly used to measure difference between probability distributions. 

Some limitations and choices were required. As mentioned, we measured distinctiveness only within the context of a single work (even for authors with multiple works). To expand beyond single works would produce very mismatched sample sizes, since some authors were prolific and some produced just one play; even with non-parametric methods, hugely mismatched sample sizes are problematic. Further, the plays span four languages and roughly five centuries, making the `distant' context seem ridiculous. As well as the selected distinctiveness statistic (median energy distance) we also recorded a `baseline’ distinctiveness, being each character’s distance from themselves. The theoretical baseline is, of course, zero, but the sample baselines will not be, so this gives us an idea of the inherent variance of the samples. Finally, when selecting characters to examine, we chose a minimum size of 2000 words. Sample sizes are somewhat arbitrary, and are matters of debate \parencite{eder_does_2015,eder_short_2017}, but this seemed a reasonable, or perhaps even slightly aggressive, lower bound.

\subsection{Area under keywords}

Our second, supplementary approach was informed by `unmasking' techniques, often employed in stylometric research \parencite{koppel_authorship_2004,kestemont_authenticating_2016,plechac_applications_2021}. Unmasking refers to a range of methods that share one goal: to measure and compare the \emph{depth} of the differences between two sets of texts. For example, an author might write both high fantasy fiction and historical novels: a classifier would have little difficulty distinguishing one genre from another by simply using superficial features (e.g. `dragons', `magic', `elves'). However, by assumption, if these most distinctive features are removed, the classifier will have more trouble determining which text came from which pool, because the texts share one deep similarity---a common authorial style. Conversely, if we compare books by two different fiction writers, these texts will also have superficial differences. However, while removing more and more distinctive features, the classifier should remain confident in distinguishing the authors from each other, because the texts do not share an authorial style that is deeply rooted in common linguistic elements and distributed over many features. By comparing the speed with which the rates of accuracy decay we can approach authorship verification problems, i.e. how plausible is that this text belongs to author A? 

We applied the same thinking to fictional characters, as opposed to authors: the distinctiveness of a character may rely on a small number of catch-phrases (`Gadzooks!' or `Cowabunga!'), or it may be driven by non-stylistic, referential factors (Mary, speaking to John is not likely to use word `Mary', but likely to use word `John', and vice-versa). On the other hand, there are characters whose speech systematically differs from the neutral language: such as when the author imitates dialects, slang, regionalism, speech and phonetic idiosyncrasies. In the former case, an imaginary classifier should quickly lose accuracy (since John and Mary speak quite similarly), but in the latter case the removal of a small number of features would not be enough to disrupt classification. 

In our case, it was impractical to use `standard', supervised (i.e. classifier-based) unmasking because individual characters, as samples, were simply too small. Instead we used word keyness---a character's relative preference for a word in the context of a given drama---to calculate an alternative distinctiveness score together with a bag of easily interpretable features per character. First, we use weighted log-odds \parencite{monroe_fightin_2008} to calculate keywords for a character relative to the speech pool of the rest of the cast; second, we represented each character by their 100 words with highest keyness, arranged by rank; finally, we measured the area under this curve, which we interpret as distinctiveness---characters with just a few key words will exhibit less area under the keyness curve. By comparing these final areas, we can compare the amount  of difference each character has in relation to all other speech in the play. In a similar manner to the bootstrapped approach, we upsample each character's word pool to match the size of the rest of the words in the play to minimise, as much as possible, the effect of sample size. 

\section{Results}

Overall, the distinctiveness energy statistic appears useful. The baseline (character vs self) is quite stable cross-linguistically, although it is slightly higher for characters with a very large share of dialogue (Fig. \ref{fig:distinctiveness_dialog}). Note also that the distinctiveness statistic appears roughly Gaussian (see Appendix \ref{app:linreg} for more discussion) and its range is relatively consistent between languages (peaking at roughly 0.20), although this consistency does not apply at the level of authors. The obvious issue is that there is a strong negative correlation between character size and distinctiveness, but this is not only a limitation of the method---lead characters naturally set the dominant style of a text (and, possibly, inherit more of the `true' authorial voice). Importantly, distinctiveness does not increase with the number of speakers in a play.  The method works best when there are reasonable sample sizes for both the examined character and the `other' class. This is illustrated by the `U' curve visible in the French corpus in Figure \ref{fig:distinctiveness_dialog} as the examined characters' dialogue share passes 50\%. As hoped, the energy-distance method does appear to capture characters who are written with distinctive idiolects, representing things like foreign accents or social class. For a discussion of this see Section \ref{sec:discussion}.

As seen in Figure \ref{fig:distinctiveness_year}, there is no clear correlation between the date of composition and character distinctiveness which suggests that language change does not disturb the measure. The finding that seems clear is that women are written differently to men. Female characters are generally more distinctive in all corpora (Fig. \ref{fig:distinctiveness_gender}), although this is not visible using the keyness AUC measure---leading us to conclude that the keyness measure has lower power. This difference in the distinctiveness of female characters can partly be explained by the fact that they tend to have smaller parts (Fig \ref{fig:dialog_gender}), and smaller characters in general are more distinctive (Fig. \ref{fig:distinctiveness_dialog}), but that is not the whole story. Female parts have more restricted 3-gram vocabularies (Fig. \ref{fig:vocab_gender}), suggesting that they are also restricted in their semantic fields. This becomes clearer when the relative frequencies of their (word) vocabularies are examined. As well as the stereotypical tendencies (women say `love', men say `sword'), the female characters, cross-linguistically, seem to be less likely to reference the `external world' of the drama. As seen in Appendix \ref{app:more_frequent}, relatively more frequent words for women are dominated by personal pronouns representing `I', `me', `you', etc. or words relating to family. The male lists are dominated by indicative articles and political terms (`law', `noble', `king', etc.).

The higher distinctiveness of female characters is further supported by a formal linear model: we fit a Bayesian multiple regression where distinctiveness was conditioned on both gender and size (characters' percentage of total dialogue). A direct gender effect is present in all corpora, as expected from Figure \ref{fig:dialog_gender}, but, when we account for variation among authors, the effect may be less pronounced than it appears (for analysis and more detailed discussed, the posterior estimates are described in Appendix \ref{app:linreg}). Our finding interlocks with the observation by \textcite{underwood_transformation_2018} that female characters found in English 18--20th century fiction displayed high distinctiveness due to the particular way they were narrated, suggesting a pervasive authorial mentality.

\section{Discussion}
\label{sec:discussion}

The measures of stylistic character distinctiveness that were proposed in this paper appear to be effective in capturing a \emph{degree} to which characters stand out from others. The most distinctive characters, by both of our metrics, often have systematically different speech, in the form of dialects, regionalisms or class markers. For example, Shakespeare's Captain Fluellen (\emph{Henry V}) is Welsh, and his accent is written for comedic effect. The systematic replacements b$\rightarrow$p and d$\rightarrow$t make him the most distinctive Shakespearean character according to both the 3-gram and word measures:

\begin{changemargin}{2cm}{2cm}
\footnotesize
\begin{drama}
\speaker{Fluellen}\phantom{x}\\ 
Your grandfather of famous memory, an't please your \\
majesty, and your great-uncle Edward the Plack \\
Prince of Wales, as I have read in the chronicles, \\
fought a most prave pattle here in France.
\speaker{King Henry V}\phantom{x}\\
They did, Fluellen.
\speaker{Fluellen}\phantom{x}\\
Your majesty says very true: if your majesties is \\
remembered of it, the Welshmen did good service in a \\
garden where leeks did grow, wearing leeks in their \\
Monmouth caps; which, your majesty know, to this \\
hour is an honourable badge of the service; and I do \\
believe your majesty takes no scorn to wear the leek \\
upon Saint Tavy's day.
\end{drama}
\end{changemargin}

Regional differences also contribute to high distinctiveness in the German corpus. For example, Emerike, written by Johanna von Weißenthurn, uses -ey instead of -ei (zwey, bey, Freylich) which is a form indicative of pre-standardised Southern German spelling. John, in Hauptmann's \emph{Die Ratten}, speaks Plattdeutsch, a variant heavily influenced by Dutch, e.g. `Det hat er jesacht, det ick noch ma hin müßte und janz jenau anjeben'.

In the French corpus, the most distinctive character by keyness is Gareau, from \emph{Le Pédant Joué} (Cyrano de Bergerac), who speaks a `patois' or rural dialect. In his critical edition, Frédéric Lachèvre comments on this distinct idiolect when Gareau is first introduced \parencite[25]{cyrano_de_bergerac_les_1921}:

\begin{changemargin}{2cm}{2cm}
\footnotesize
Cyrano a fabriqué de toutes pièces le patois de Gareau. Le manuscrit de la BN donne un langage tout différent que celui imprimé en 1654, la pronociation des mots n'est pas tout à fait la même. Nous avons naturellement maintenu pour Gareau le texte de 1654 publié par Cyrano lui-même.\\

Cyrano created the patois of Gareau from scratch. The manuscript of the [Bibliothèque Nationale] offers quite a different language to the one printed in 1654, the pronunciation of the words is not quite the same. We have naturally maintained for Gareau the text of 1654 published by Cyrano himself.
\end{changemargin}

The most distinctive Russian characters come from Ostrovskii, who gave the main stage to Muscovite merchants and their families with their vernacular, non-aristocratic language. Tolstoy's Nikita (high on both the 3-gram and keyness lists) from \emph{The Power of Darkness} has heavily stylised speech suggestive of Western or Southern Russian dialects, e.g. featuring a word-initial [w].

It must be borne in mind, however, that dialects or accents do not automatically cause high distinctiveness---what is being detected is the \emph{difference} in speech patterns. In a text where everyone speaks Welsh, an English character would score highly on distinctiveness, and vice versa. Cross-linguistic inference must also account for systematic language differences: the lexical and morphological features of the various languages lead naturally to different probability distributions for both words and $N$-grams (although the exact nature of those differences is too complex to grapple with here). Word-based distinctiveness measures permit easier interpretation, but appear less (statistically) powerful. In addition, word-based measures operate in much higher dimensions, with all the usual problems that entails (sparsity, the `curse of dimensionality', etc. See, for example \textcite{moisl_finding_2011}). Finally, word-based measures naturally invite lemmatisation for highly inflected languages (like Russian and German), which might cause problems for future work dealing with languages that are non-standard, historical, or otherwise less well-resourced.

We have noted that our distinctiveness measure has a strong negative correlation to the size of the character. This relationship should not be understood as a simple artefact that renders our measurement useless.  Distinctive speech is always a construct, a subset of linguistic and stylistic reality. If a minor character has just a few lines about gallows and graves---like Shakespeare's gravedigger---we will never know more about their language. However, \emph{Hamlet} is not \emph{only} about gallows and graves; if we imagine bootstrapping the gravedigger's speech, it would be endlessly populated by these few words: we don't know how the gravedigger would speak when ruling a country, or murdering their uncle. From this perspective, a protagonist is more likely to represent lexical and stylistic norm, while minor characters will sample the Other in their ethnic, dialectal, or professional distinctiveness.

Despite the few limitations, we hope that these measures of character distinctiveness will support improved theories about style, characterisation and history. The most important question to be asked concerns the source(s) of this representational distinctiveness that authors instil in their characters. To even begin to address this issue, we need much richer annotation for characters: their social class, profession, region of origins, age. Determining the drivers of distinctiveness will not be easy. Even to carefully verify the effect of character gender was quite complicated. We know that part of the effect comes from size: women are more likely to be minor characters. However, it is reasonable to assume that gender difference can also be confounded by genre (e.g. in comedies there are more women playing larger roles) and social class (rural people speak more in comedies). There is also the effect of time: changing the relative dynamics of character sizes \parencite{algee-hewitt_distributed_2017}, improving the representation of women as dramatists and altering the depiction of social class---all of which complicates the analysis even further. However, having a clear summary measure for a character's stylistic distinctiveness may help us to refine our theories about the speech of fictional characters, leading in turn to better causal models.

\section{Availability of Data and Code}

The details of our approach, including data acquisition and preprocessing, are published in a Zenodo repository, allowing for full replication of all reported results: \url{https://doi.org/10.5281/zenodo.7383687}.

\section*{Acknowledgements}
AŠ, JB, LHL and ME were funded  by the ``Large-Scale Text Analysis and Methodological Foundations of Computational Stylistics'' project (SONATA-BIS \\2017/26/E/HS2/01019). BN is also grateful to QuaDramA, funded by Volkswagen Foundation and the DFG-priority programme Computational Literary Studies, for financing the presentation of the paper at the workshop.

\printbibliography
\newpage
\begin{appendices}
\pagenumbering{gobble}

\section{Relatively-More-Frequent Words}
\label{app:more_frequent}
\begin{table}[htbp]
\centering
\footnotesize
\begin{tabular}{llllll}
 \multicolumn{2}{c}{\normalsize French} & \multicolumn{2}{c}{\normalsize German} & \multicolumn{2}{c}{\normalsize Shakespeare} \\
  \multicolumn{1}{c}{Female} &      \multicolumn{1}{c}{Male} & 
  \multicolumn{1}{c}{Female} &      \multicolumn{1}{c}{Male} &
  \multicolumn{1}{c}{Female} &      \multicolumn{1}{c}{Male} \\
  \cmidrule(r){1-2} \cmidrule(rl){3-4} \cmidrule(l){5-6} \\
        vous &    diable &       ach &       der &     husband &       the \\
       époux &        la &         o &       die &         you &        of \\
        mère &       ami &        du &    teufel &        alas &      this \\
       amant &       les &     vater &       und &        love &       sir \\
        mari &   parbleu &    mutter &       ein &    husbands &       and \\
       tante &    maître &        er &       des &          me &        we \\
       hélas &   morbleu &      mich &        in &       romeo &      king \\
       coeur &       des &     liebe &       den &    lysander &       our \\
      rivale &      amis &      mama &      kerl &      willow &     their \\
          ne &    morgué &      papa &    kaiser &     pisanio &      duke \\
 malheureuse & serviteur &      nein &       ihr &      sister &     three \\
        quil &     belle &       dat &      euch &     nerissa &       her \\
         mon &       vin &      mein &       auf &       yours &        to \\
          me &        un &      herz &       dem &           o &      whom \\
       maman &   heureux &    gemahl &       wir &        pray &  lordship \\
       frère &      leur &      gott &     könig &      mother &        in \\
      fâchée &      rome & geliebter &     sache &       nurse &     stand \\
        père &    peuple &      kind &      also &           i &     noble \\
     obligée &     boire &       ihn &        hm &       woman &        ha \\
        sûre &   soldats &    lieber &  majestät &    malvolio &       dog \\
     dorante &     peste &     nicht &      oder &     prithee &   certain \\
          il &      prêt &      nich &      volk &          my &      kate \\
       soeur &     rival &       sie &      euer &     orlando &    master \\
       amour &        dé &      mann &       das &       boyet &     sword \\
         lui &     sénat &       weh &     unter &          do &    follow \\
    heureuse &        ça &       dir &        im &       false &  soldiers \\
         que & messieurs &      dich &       zum &        ring &       his \\
      pleurs &    coquin & mellefont &    freund &      emilia &    caesar \\
       cruel &      gens &        ja &     krieg &      refuse &        us \\
      lamour &        du &       ihm &     durch &     troilus &       law \\
   chevalier &    allons &     angst &     hölle &     pilgrim &   friends \\
       seule &    beauté &  freundin &        zu &     windsor &      york \\
       aimée &        au &       wat &    gnaden &       would &     money \\
     lingrat &    lhomme &      doch &      wein &    rosalind &    pompey \\
      valère &       par &  mamachen &      heer &        such &   england \\
        aime &    obligé &        so &       mit &        weep &   present \\
       aimer &        lé &       mir &    bürger &       faith &   warwick \\
       maime &       bon &     fritz &     jeder &        suit &     great \\
        hans &     cents &      arme &    herren &          am &     heads \\
       chère &     bâton &     gurli &       rom &       diana &     ready \\
      ingrat &    quatre &      lieb &      land &       never &  business \\
\end{tabular}
\caption{40 most relatively-more-frequent words (Weighted Log-Odds) for the French, German and Shakespearean corpora.}
\label{tab:most_common}
\end{table}
\section{Bayesian Regression Models: effect of gender on distinctiveness}
\label{app:linreg}

\begin{figure}[b]
  \centering
  \includegraphics[width=0.95\textwidth]{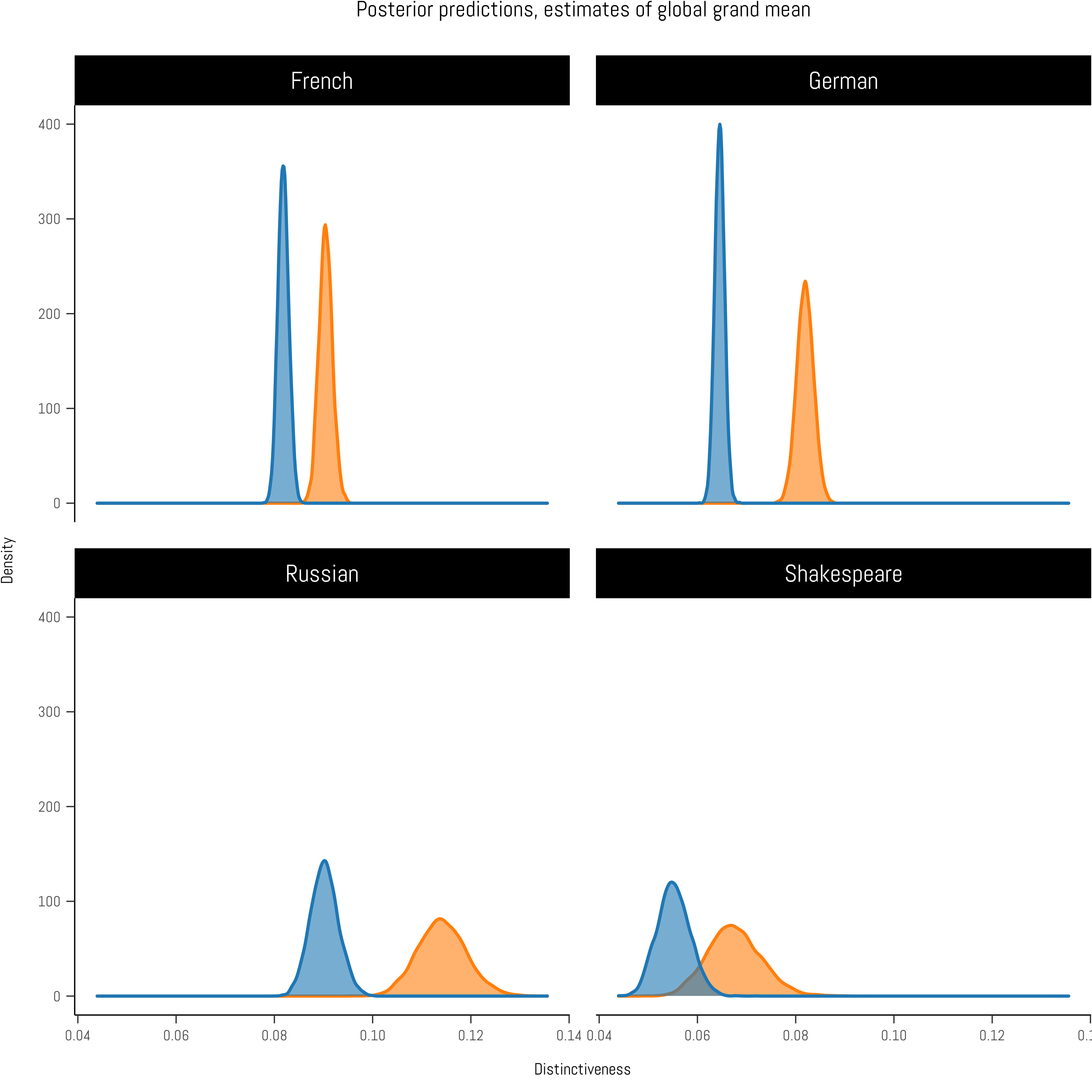}
  \caption{Character distinctiveness, predicted from posterior,  estimate of grand mean (no group-level effects), 6000 draws. Predictions are made for a counterfactual "median" character role, who has 20.9\% of dialogue share. Predictions are presented at natural scale.}
  \label{fig:posterior_grand}
\end{figure}

\begin{figure}
  \centering
  \includegraphics[width=0.95\textwidth]{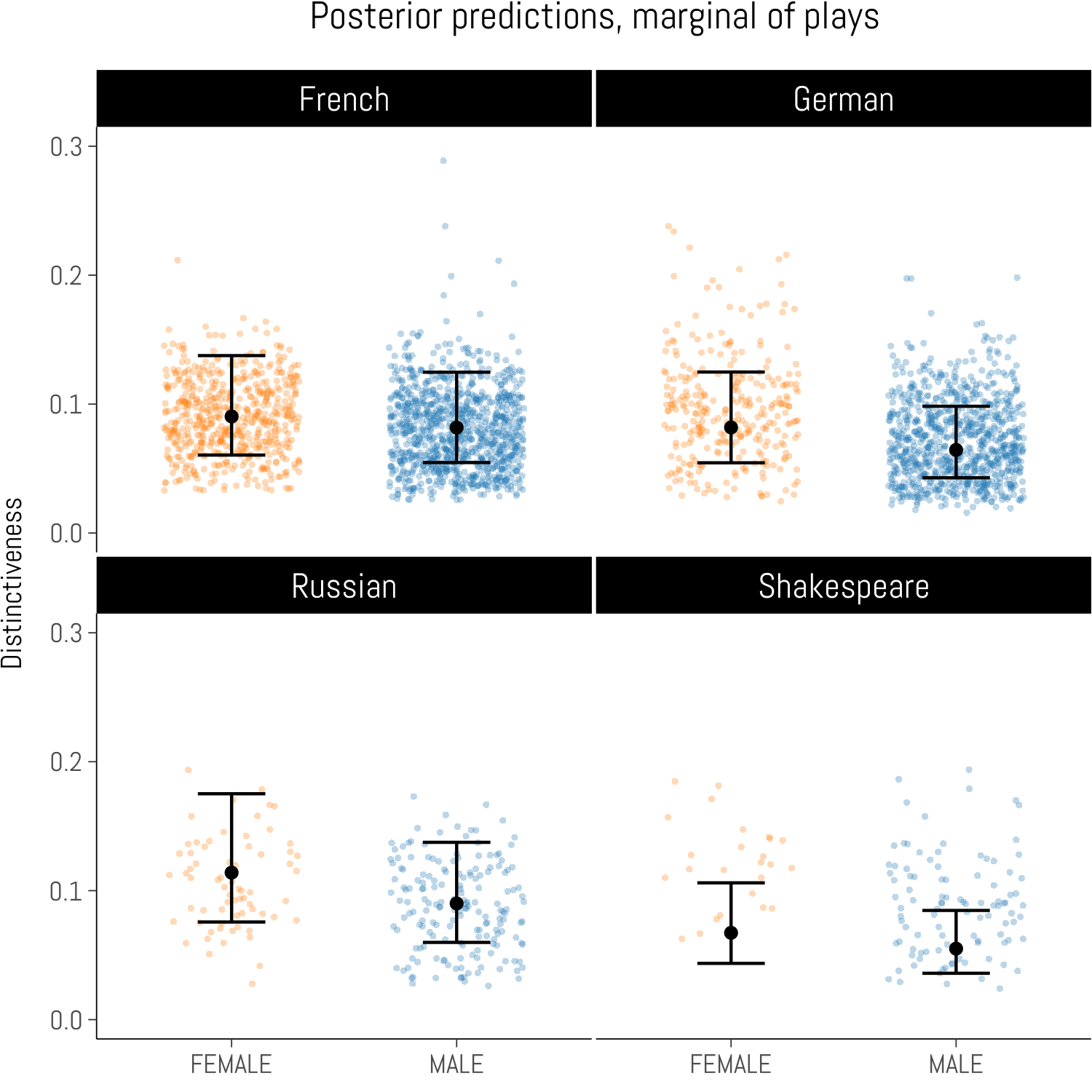}
  \caption{Posterior predictions for gender, marginal of individual plays. Errorbars show .95 CI. Empirical data is plotted in colour, 5 extreme cases ($>$0.3) are filtered out. Predictions are presented at natural scale.}
  \label{fig:posterior_marginal}
\end{figure}

Is the perceived gender effect `real'? In technical terms, what is the direct influence of character gender (G) on distinctiveness  scores (D) across traditions (T), conditioned on the share of dialogue they have (S)? To answer this, we fit a Bayesian multilevel multiple regression with group-level estimates for individual plays (P). We chose to model at the level of plays both because our D statistic is tied to the context of a single play, and because character features coming from the same play are not independent (e.g. there cannot be two characters with 60\% of the dialogue). Modelling this way also significantly improved predictions. Gender is allowed to interact by corpus, yielding a single, cross-linguistic model that makes compatible predictions for different traditions.  In brms formula syntax:

\texttt{log(D) $\sim$ G * T + T*(S + I(S\string^2)) + (1|P)}

Based on sample observation, we used a Gaussian prior for log-transformed D scores. We could have also fitted the original values, but D scores have extreme outliers that extend the tail: the model has much easier time with sampling and chain convergence on a log-transformed domain. We chose a quadratic term for S, because the relationship between D and S is U-shaped. Importantly, `unknown' gender entities are filtered, because often (but not always) this is not data that is missing, but entries that are incompatible with a binary classification:%
\footnote{In modern terms, it is vexing to be forced to reduce characters to a  gender binary, but since gender non-conforming characters are virtually unrepresented in this predominantly historical corpus, the point is moot.
}
primarily collective or compound entities (people, choirs, soldiers). It would have been possible to use standard strategies, like imputation, to `repair' the data, but that approach would be incorrect.

Posterior estimates for distinctiveness by gender are shown in Figure \ref{fig:posterior_grand}. Based on the figure, we can be most confident about the difference in German and least confident in Shakespeare (few characters and, specifically, few women with large dialogue shares). The differences in means, however, appear consistent. As calculated from the posterior: in French, female characters are more distinctive by only .009 ($\pm$ .003); in German, by .017 ($\pm$ .003); in Russian by .023 ($\pm$ .009, the widest CI); and in Shakespeare by .012 ($\pm$ .008).

To understand the full extent of variation across different plays, it is useful to look at the marginal posterior means of the plays (Fig. \ref{fig:posterior_marginal}). Here, the difference in distinctiveness between genders remains visible, but there is a better estimation of the global uncertainty and variation across different texts. Note that the confidence intervals in Fig. \ref{fig:posterior_marginal} are asymmetric (wider on the upper arm), having been transformed from symmetric intervals on a log domain.

\end{appendices}

\end{document}